\documentclass[sigconf]{acmart}
\renewcommand\footnotetextcopyrightpermission[1]{} 
\usepackage{booktabs} 
\usepackage{subfigure} 
\usepackage{rotating}
\usepackage[normalem]{ulem}
\usepackage{verbatim}
\usepackage{verbatimbox}
\usepackage{url}

\usepackage{amsmath}

\setcopyright{rightsretained}

\newcommand{\mypartop}[1]{\vspace{0mm}{\noindent\textbf{#1}}}
\newcommand{\mypar}[1]{\vspace{0mm}{\noindent\textbf{#1}}}

\newif\ifdraft
\draftfalse
\ifdraft
  \newcommand{\misha}[1]{{\color{blue}{M: #1}}}
  \newcommand{\jasper}[1]{{\color{magenta}{J: #1}}}  
  \newcommand{\vitto}[1]{{\color{red}{V: #1}}}
  \newcommand{\rodrigo}[1]{{\color{violet}{R: #1}}}
\else
  \newcommand{\misha}[1]{}
  \newcommand{\jasper}[1]{}
  \newcommand{\vitto}[1]{}
  \newcommand{\rodrigo}[1]{}
\fi

\newcommand{\initempty}{\textit{init-empty~}}
\newcommand{\initauto}{\textit{init-auto~}}
\newcommand{\initautonms}{\textit{init-auto+nms0.5~}}

\newcommand{\initautonmsSSN}{\textit{init-auto+nms0.5+sortscore-topN~}}
\newcommand{\initautonmsSDN}{\textit{init-auto+nms0.5+sortdistance-topN~}}

\newcommand{\initautonmsSDT}{\textit{init-auto+nms0.5+sortdistance-top3~}}

\newcommand{\mact}{micro-actions~}

\acmDOI{XXX}

\acmISBN{123-4567-24-567/08/06}

\acmConference[ACMMM'18]{ACM Multimedia}{2018}{Seoul, Korea}
\acmYear{2018}
\copyrightyear{2018}

\acmPrice{15.00}

\begin{document}
\title{Fluid Annotation: A Human-Machine Collaboration Interface for Full Image Annotation}


\author{Mykhaylo Andriluka$^{*}$\qquad Jasper R. R. Uijlings$^{*}$\qquad Vittorio Ferrari\vspace{0.1cm}}
\affiliation{%
  \institution{Google Research}
  \city{Z\"urich}
  \state{Switzerland\vspace{0.5cm}}
}




\renewcommand{\shortauthors}{M Andriluka, J. R. R. Uijlings and V. Ferrari}

\begin{abstract}
We introduce Fluid Annotation, an intuitive human-machine collaboration
interface for annotating the class label and outline of every object and
background region in an image\footnote{Live demo of the interface is available at \url{fluidann.appspot.com}}.
Fluid annotation is based on three principles:
\mbox{\emph{(I) Strong Machine-Learning aid.}}
We start from the output of a strong neural network model, which the annotator can edit by correcting the labels of existing regions, adding new regions to cover missing objects, and removing incorrect regions. The edit operations are also assisted by the model.
%
\mbox{\emph{(II) Full image annotation in a single pass.}}
As opposed to performing a series of small annotation tasks in isolation~\cite{lin14eccv,russakovsky15ijcv}, we propose a unified interface for full image annotation in a single pass.
\mbox{\emph{(III) Empower the annotator.}}
We empower the annotator to choose what to annotate and in which order. This
enables concentrating on what the machine does not already know, i.e. putting
human effort only on the errors it made. This helps using the annotation budget
effectively. Through extensive experiments on the COCO+Stuff dataset~\cite{lin14eccv,caesar18cvpr}, we demonstrate that Fluid Annotation leads to accurate annotations very efficiently, taking $3\times$ less annotation time than the popular LabelMe interface~\cite{russel08ijcv}.

\end{abstract}

%
%



\maketitle
\renewcommand*{\thefootnote}{\fnsymbol{footnote}}
\footnotetext{The authors contributed equally.}
\renewcommand*{\thefootnote}{\arabic{footnote}}

\section{Introduction}

The need for large amounts of high-quality training data is quickly becoming a major bottleneck in deep learning. Popular computer vision models continue to grow
in size (e.g.~\cite{krizhevsky12nips,ren15nips,simonyan15iclr,szegedy17aaai,he16cvpr,he17iccv}) and larger amounts of
data continues to improve accuracy~\cite{caesar18cvpr,huh16nips,mahajan18arxiv,schroff15cvpr,sun17iccv}).
Annotation is especially expensive for models requiring training images annotated with the class label and outline of every object and background region~\cite{chen18pami,liu16iclr,long15cvpr,wu16arxiv}.
For example, annotating one image of the COCO dataset~\cite{lin14eccv} required 80 seconds per object to draw a polygon on it~\cite{lin14eccv},
plus 3 minutes annotating background regions~\cite{caesar18cvpr}, for a total of 19 minutes on average.
Similarly, fully annotating one image of the Cityscapes dataset~\cite{cordts16cvpr} took 1.5 hours.

\begin{figure}[t]
    \centering
    \begin{tabular}{c}
      \includegraphics[width=0.85\columnwidth]{figures/image-000130-115885.png}\\
      \includegraphics[width=0.85\columnwidth]{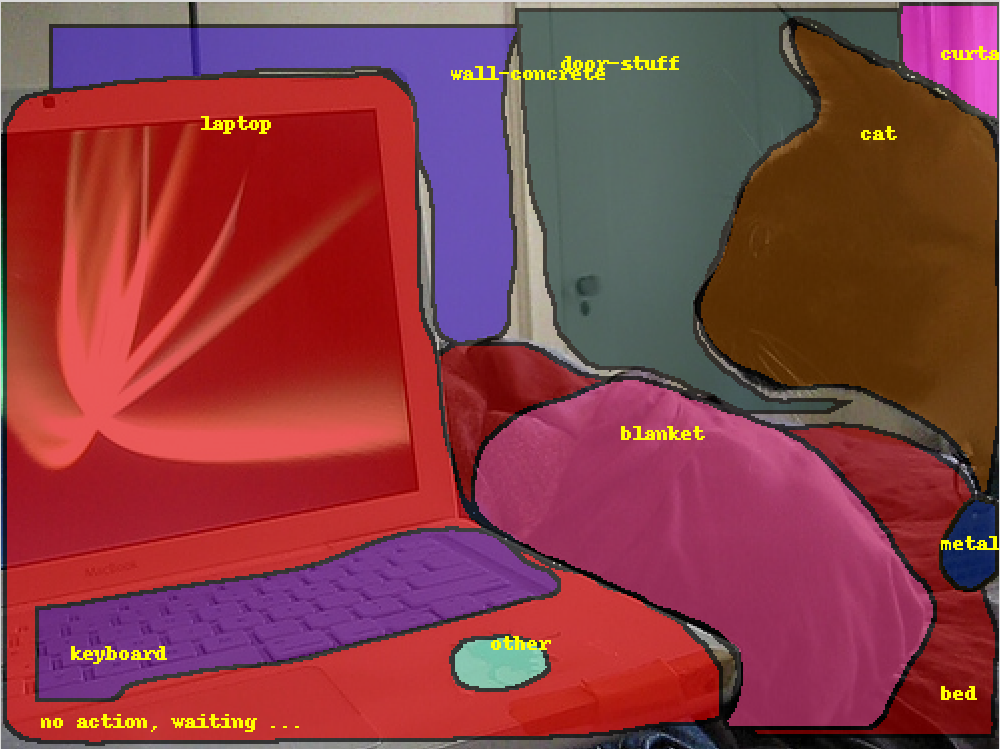}\\

    \end{tabular}
    \caption{Example of an annotation result: original image (top) and full annotation of objects and background obtained with just 9 annotation actions using our approach (bottom).}
    \label{fig:teaser}
\end{figure}

In this paper we propose Fluid Annotation, a new human-machine collaboration interface for annotating all objects and background regions in an image.
The goal is to have a very efficient and natural interface, which can produce high quality annotations with much less human effort 
than traditional manual interfaces (Fig.~\ref{fig:teaser}).

Fluid Annotation is based on three design principles:

\paragraph{(I) Strong Machine-Learning aid.}
Popular semantic segmentation
datasets~\cite{caesar18cvpr,cordts16cvpr,everingham15ijcv,lin14eccv,mottaghi13cvpr,zhou17cvpr} are
annotated fully manually which is very costly.
Instead Fluid Annotation starts from the output of a neural network model~\cite{he17iccv}, which the annotator can edit by
correcting the label of existing regions,
adding new regions to cover missing objects,
and removing incorrect regions (Fig.~\ref{fig:actionexample}).
By starting from the most likely machine-generated output,
and supporting quick and intuitive editing operations, 
Fluid Annotation enables full image annotation in a short amount of time.

\paragraph{(II) Unified interface for full image annotation in a single pass.}
Many datasets are annotated using a
series of micro-tasks such as indicating object presence in an image~\cite{lin14eccv,russakovsky15ijcv,deng14chi}, clicking on instances of a
specific class~\cite{lin14eccv}, or drawing a polygon or a box around a single instance~\cite{su12aaai,everingham15ijcv,lin14eccv}.
Correspondingly, previous ML-aided interfaces focus on a single micro-task, such as segmenting individual objects~\cite{acuna18cvpr,boykov01iccv,Rother04-tdfixed,castrejon17cvpr,jain13iccv,jain16hcomp,liew17iccv,nagaraja15iccv,xu16cvpr}
or annotating bounding boxes~\cite{papadopoulos16cvpr},
or they focus on selecting which micro-task to assign to the annotator~\cite{konyushkova18cvpr,russakovsky15cvpr,vijayanarasimhan09cvpr}. 
In contrast, with Fluid Annotation we propose a
single, unified ML-aided interface to do full image annotation in a single pass.

\paragraph{(III) Empower the annotator.}
In most annotation approaches there is a fixed sequence of annotation
actions~\cite{caesar18cvpr,cordts16cvpr,everingham15ijcv,lin14eccv,mottaghi13cvpr,zhou17cvpr} or the
sequence is determined by the machine~\cite{konyushkova18cvpr,russakovsky15cvpr,vijayanarasimhan09cvpr}.
In contrast, Fluid Annotation empowers the annotator:
he sees at a glance the best available machine segmentation of all scene elements, and then decides what to annotate and in which order.
This enables to focus on what the machine {\em does not already know}, i.e. putting human effort only on the errors it made, and typically addressing the biggest errors first. 
This helps using the annotation budget effectively 
, and also steers towards labeling hard examples first. Focusing on hard examples is known to beneficial to improve the model later on (e.g.~\cite{felzenszwalb10pami,freund97jcss,shrivastava16cvpr}).

Our contributions are:
(1) We introduce Fluid Annotation, an intuitive human-machine collaboration interface for fully annotating an image in a single pass.
(2) By using simulated annotators, we demonstrate the validity of our approach and optimize the effectiveness of our interface.
(3) Using expert human annotators, we compare our Fluid Annotation interface with the popular LabelMe interface~\cite{russel08ijcv} and demonstrate that we can produce annotations of similar quality while reducing time by a factor of $3\times$.

\section{Related Work}

\mypar{Weak supervision.}
A common approach to reduce annotation effort is to use weakly labeled data. For example, several
works train object class detectors from image-level labels only (i.e. without annotated bounding boxes)~\cite{bilen16cvpr,cinbis14cvpr,deselaers10eccv,haussmann17cvpr,kantorov16eccv,zhu17iccv}.
Other works require clicking on a single point per object in images~\cite{papadopoulos17cvpr}, or per action in video~\cite{mettes16eccv}.
Semantic segmentation models have been trained using image-level labels
only~\cite{kolesnikov16eccv,pathak15iccv}, using
point-clicks~\cite{bearman16eccv,bell15cvpr,wang14cviu}, from
boxes~\cite{khoreva17cvpr,maninis18cvpr,papadopoulos17iccv} and from
scribbles~\cite{lin16cvpr,xu15cvpr}.

A recent variant of weakly supervised learning is the so-called "webly supervised learning", where one learns from large amounts of noisy data crawled from the web~\cite{berg06cvpr,fergus10ieee,li06acmmm,jin17cvpr,li17cvpr}. While large amounts of images with image-level labels can be obtained in this manner, full-image segmentation annotations cannot be readily crawled from the web.

\mypartop{Human-machine collaborative annotation.}
Several works have explored interactive annotation, where the human annotator and the machine model collaborate. In weakly supervised works the human provides annotations only once before the machine starts processing. Interactive annotation systems instead iterate between humans providing annotations and the machine refining its output. 

Most works on interactive annotation focus on a single, very specific task. In particular, many
works address segmenting a single object instance by combining a machine model and user input within an interactive
framework~\cite{acuna18cvpr,boykov01iccv,castrejon17cvpr,Rother04-tdfixed,jain13iccv,jain16hcomp,liew17iccv,nagaraja15iccv,xu16cvpr}.
Typically, the machine first predicts an initial segmentation, which is then corrected by
clicks~\cite{jain16hcomp,liew17iccv,xu16cvpr},
scribbles~\cite{boykov01iccv,nagaraja15iccv,Rother04-tdfixed}, or by editing polygon vertices~\cite{acuna18cvpr,castrejon17cvpr}. The machine then updates the segmentation based on the user input and the process iterates until the user is satisfied.
Other works address other specific tasks, such as annotating bounding boxes of a given class known to be present in the image~\cite{papadopoulos16cvpr}, and fine-grained image
classification through attributes~\cite{branson10eccv,parkash12eccv,biswas13cvpr,wah14cvpr}.
Instead of focusing on a specific task, we propose a full image annotation interface, covering the class label and outlines of all objects and background regions.

Another research direction focuses on selecting which micro-task to assign to the annotator~\cite{russakovsky15cvpr,konyushkova18cvpr,vijayanarasimhan09cvpr}. In~\cite{konyushkova18cvpr} they train an
agent to automatically choose between asking an annotator to manually draw a bounding box or to verify a machine-generated box.
In~\cite{russakovsky15cvpr} the set of micro-tasks also includes asking for an image-level label and finding other missing instances of a class within the same image.

\mypar{Active learning.}
Active learning systems start with a partially labeled
dataset, train an initial model, and ask human annotations for examples which are expected to improve the model accuracy the most. Active learning has been used to train whole-image
classifiers~\cite{joshi09cvpr,kapoor07iccv,kovashka11iccv,qi08cvpr}, object class
detectors~\cite{vijayanarasimhan14ijcv,yao2012cvpr}, and semantic
segmentation~\cite{siddiquie10cvpr,vijayanarasimhan08nips}.
While active learning focuses on which examples to annotate, we explore creating a human-machine collaboration interface for full image annotation.

\section{Fluid Annotation}
\bgroup
\tabcolsep 0.5pt

\begin{figure*}[t]
\vspace{-0.5cm}
    \centering
    \begin{tabular}{ccc}
      \includegraphics[width=0.33\textwidth]{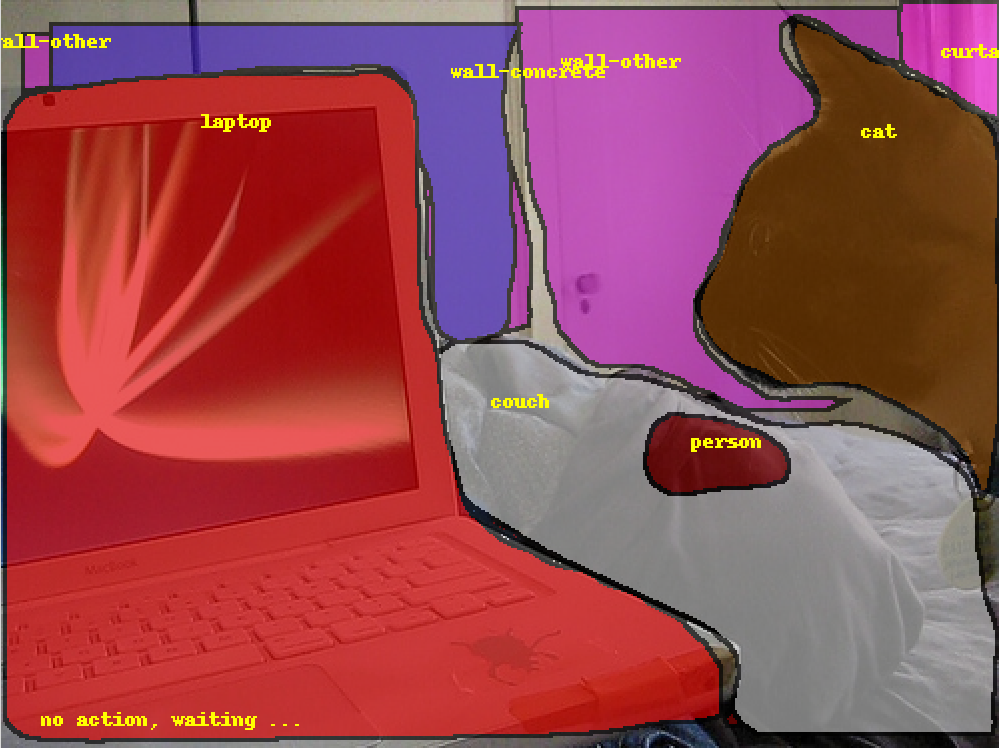}&
      \includegraphics[width=0.33\textwidth]{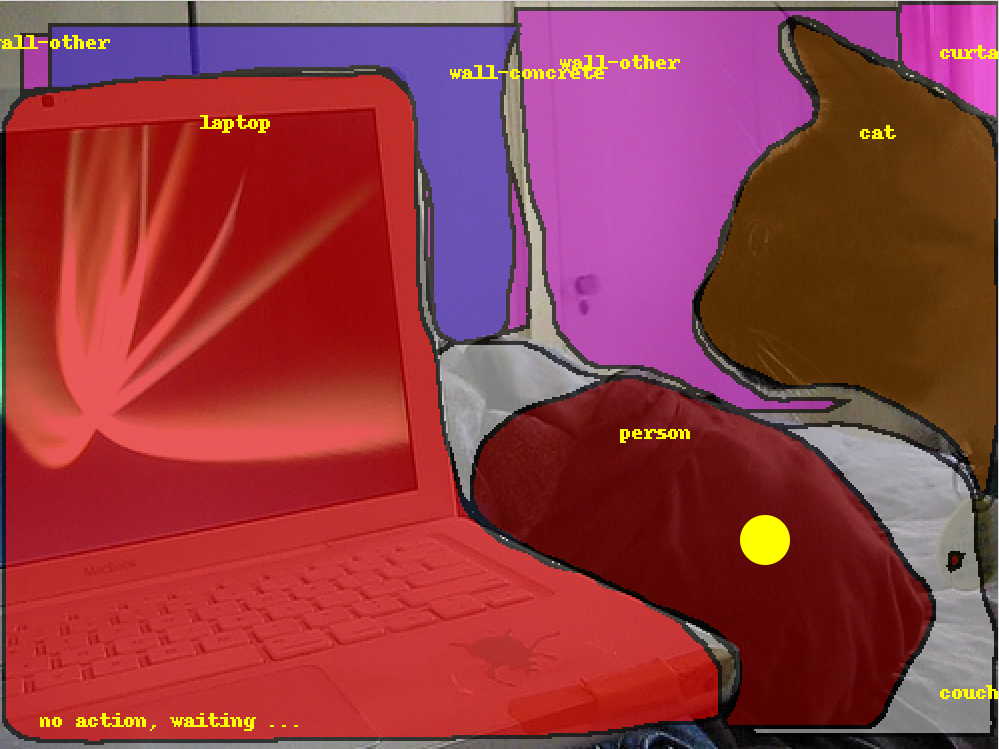}&
      \includegraphics[width=0.33\textwidth]{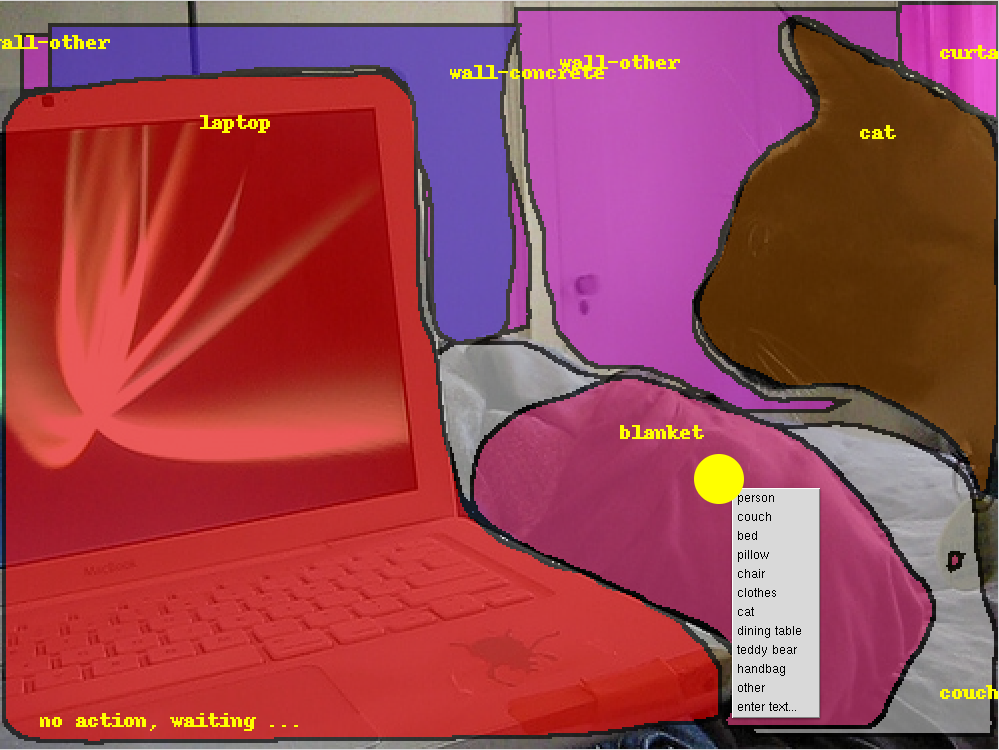}\\
      a) automatic initialization & b) step 1: ``reorder'' & c) step 2:  ``change label''\vspace{0.2cm}\\
    \end{tabular}
    \begin{tabular}{ccc}
      \includegraphics[width=0.33\textwidth]{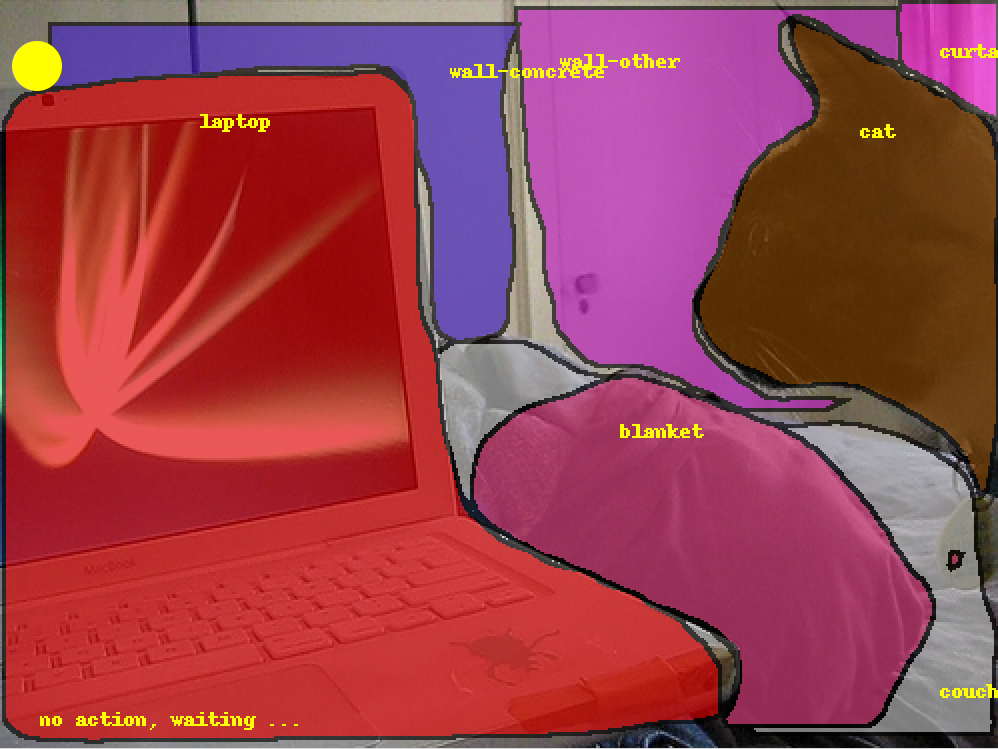}&
      \includegraphics[width=0.33\textwidth]{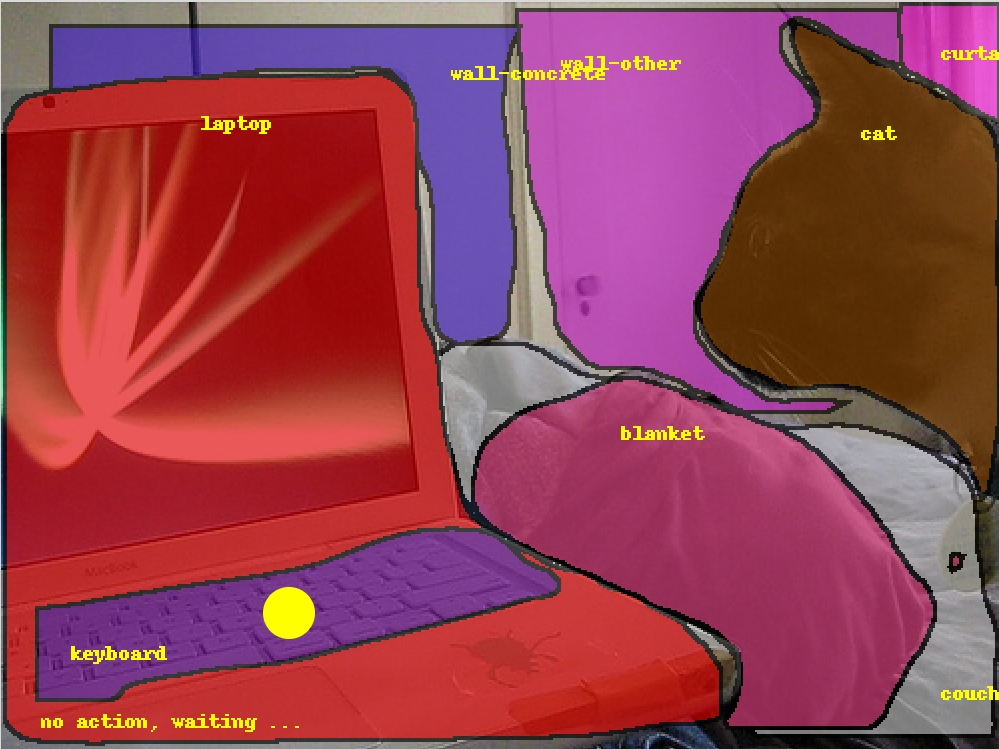}&
      \includegraphics[width=0.33\textwidth]{figures/annotool_screen_imgidx130_extralabels.png}\\
      d) step 3:  ``remove segment'' & e) step 4:  ``add segment'' & f) final result after step 9\\      
    \end{tabular}
    \caption{Example of the annotation process starting from the automatic initialization
      (top left) and progressing towards the final result (bottom right). Yellow circle marks the location of the mouse click.}
    \label{fig:actionexample}
\end{figure*}
\egroup

\subsection{The task: full image annotation}
\label{sec:task}

We address the task of full image annotation~\cite{caesar18cvpr,cordts16cvpr,kirillov18arxiv,mottaghi14cvpr}: 
We want to annotate every object and background region in the image.
The set of classes to be annotated is fixed and predefined.
We consider both \emph{thing} classes, countable objects with a well-defined shape (e.g. \emph{cat} and \emph{bus}), and \emph{stuff} classes, which are amorphous and have no distinct parts (e.g. \emph{grass} and \emph{road}). 
For thing classes, each individual object instance needs to be annotated by its class label and a region defining its spatial extent. For example, there might be 3 cats in the image, each annotated as its own separate region.
For stuff classes, all pixels needs to be annotated with their class label, but there is no concept of instances. For example, all grass pixels in an image need to be labelled as grass, but it does not matter whether they are annotated as one single big region or split into multiple regions.
An example of a fully annotated image is illustrated in Fig.~\ref{fig:teaser}.

This task definition corresponds exactly to the {\em Panoptic Segmentation} task set out by~\cite{kirillov18arxiv}. It subsumes most previous tasks in image understanding, including image classification (only image-level labels, no localization)~\cite{caltech256}, object detection (only bounding-boxes, no outlines)~\cite{everingham15ijcv,russakovsky15ijcv,openimages}, instance segmentation (only things, no stuff)~\cite{girshick14cvpr,hariharan14eccv}, semantic segmentation (no separation between different object instances of the same class)~\cite{shotton09ijcv,tighe13ijcv}.

\subsection{Overview of the Fluid Annotation interface}

Fig.~\ref{fig:actionexample} gives an overview of the fluid annotation interface and the user interactions it supports.
Given an image to be annotated (Fig.~\ref{fig:actionexample}a), we first apply a neural network model~\cite{he17iccv} to produce a large, overcomplete set of overlapping segments, aiming at covering most objects and stuff regions in the image. We call this the \emph{proposal set}.
We then create the most likely machine-generated annotation by automatically choosing a subset of the proposal set, which we call the \emph{active set}. We resolve pixel-level ambiguities by introducing a depth ordering on the active set such that each pixel is only attributed to a single segment. The active set with its depth ordering defines the initial annotation. This is presented to the user through the Fluid Annotation interface, which is deliberately kept simple and shows only the image with the annotation overlaid, as shown in Fig.~\ref{fig:actionexample}.

The annotator can edit the current annotation by carrying out a series of actions out of the following set:
(a) change the label of an active segment,
(b) change the depth order of an active segment.
(c) remove an active segment,
(d) add a segment to the active set, by selecting one out of the proposal set. 
The annotator is free to choose which actions to perform and in which order to perform them.
The resulting interface enables efficient, full image annotation: only 9 actions are needed for the example in Fig.~\ref{fig:actionexample}.

\subsection{Interface elements: segments and labels}
\label{sec:segment_soup}
  
Fluid Annotation operates on a proposal set of segments with their corresponding labels. For Fluid Annotation to work well, we have two requirements:
(I) the proposal set should cover most of the objects and stuff regions in the image. This requirement can be satisfied by operating on a large and diverse set of segments~\cite{arbelaez14cvpr,carreira10cvpr,endres14pami,uijlings13ijcv}.
(II) Each segment in the proposal set should come with a corresponding \emph{class label} and \emph{segment score}, where the segment score represents the confidence of the class label.
This requirement can be satisfied by using one of the strong modern instance segmentation models~\cite{dai16eccv,he17iccv,chen17arxiv}.

In practice we create our proposal set using Mask-RCNN~\cite{he17iccv} with Inception-Res\-Net~\cite{szegedy17aaai} using the TensorFlow implementation of~\cite{huang17cvpr}.
Mask-RCNN originally covered only thing classes. We train a second Mask-RCNN model dedicated to stuff classes by treating connected stuff regions as positive training examples (i.e. two disjoint ``grass'' regions in the same image will result in two separate training examples).
To increase the number of output segments during inference we make two small modifications: (1) increase the number
of proposals coming from the RPN from 300 to 500; (2) for the final masks, we keep the per-class
Non-Maximum-Suppression of the final boxes to $IoU > 0.6$, but increase the total number output masks from 100 to 500.
As we use two models this leads to a proposal set of size 1000 per image.

\subsection{Interface actions}
\label{sec:actions}

We now specify in more detail the editing actions available to the annotator (Fig.~\ref{fig:actionexample}).

\bgroup
\tabcolsep 0.5pt

\begin{figure}[t]
    \centering
    \begin{tabular}{c@{\hskip0.2cm}cc}
       &\textbf{sort by score} & \textbf{sort by distance}\\
      
       \addvbuffer[12ex]{\textbf{1}}&
      \includegraphics[width=0.5\columnwidth]{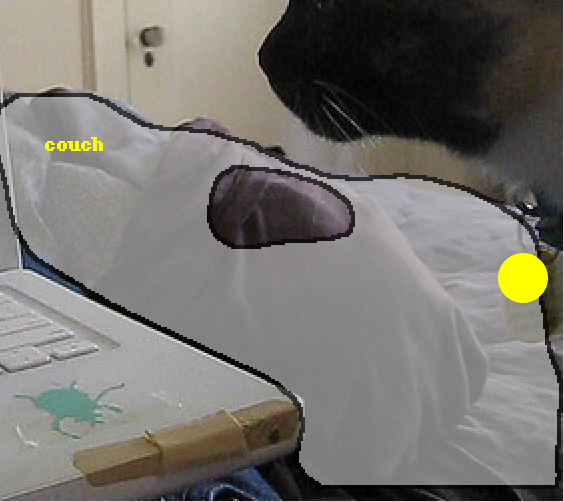}&
      \includegraphics[width=0.5\columnwidth]{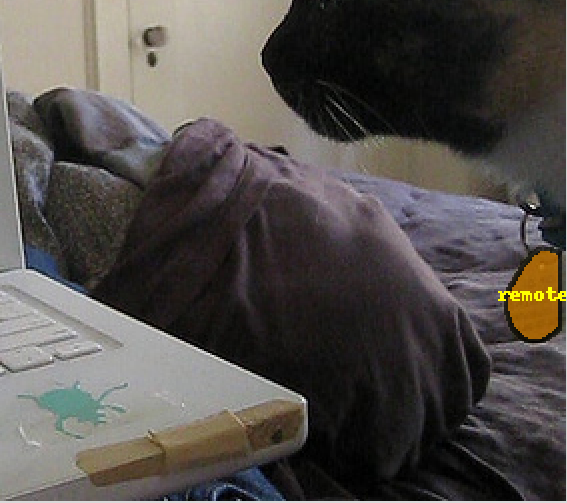}\\
      
      \addvbuffer[12ex]{\textbf{2}}&
      \includegraphics[width=0.5\columnwidth]{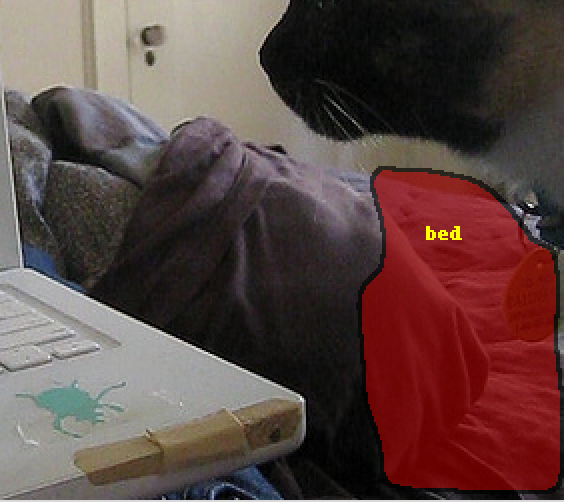}&
      \includegraphics[width=0.5\columnwidth]{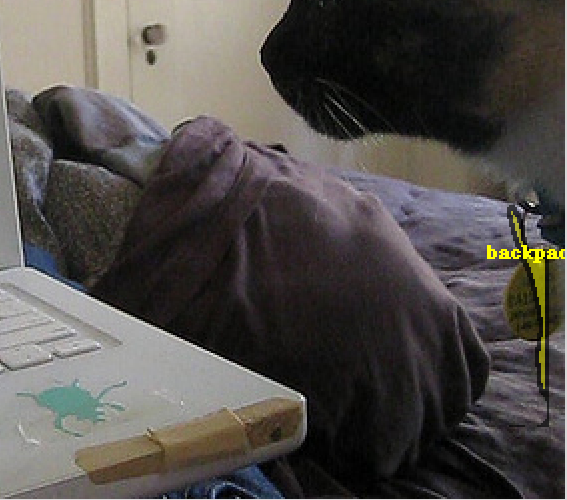}\\
      
      \addvbuffer[12ex]{\textbf{3}}& 
      \includegraphics[width=0.5\columnwidth]{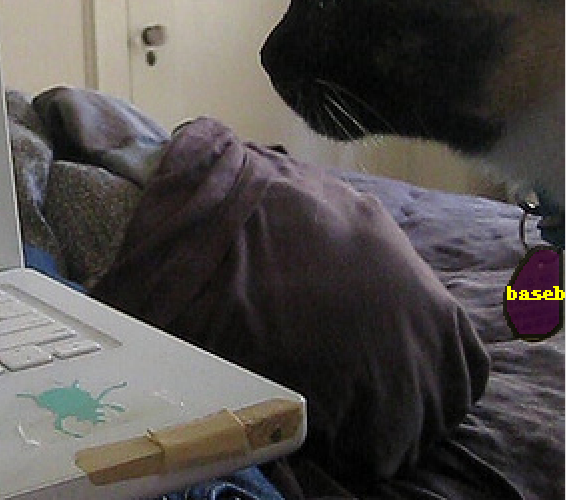}&
      \includegraphics[width=0.5\columnwidth]{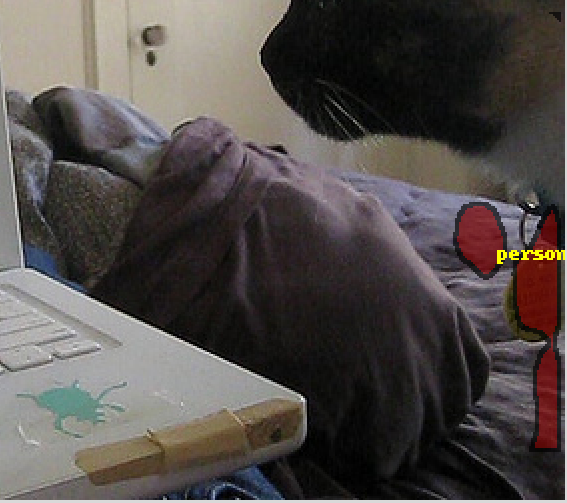}\\
      
      \addvbuffer[12ex]{\textbf{4}}&
      \includegraphics[width=0.5\columnwidth]{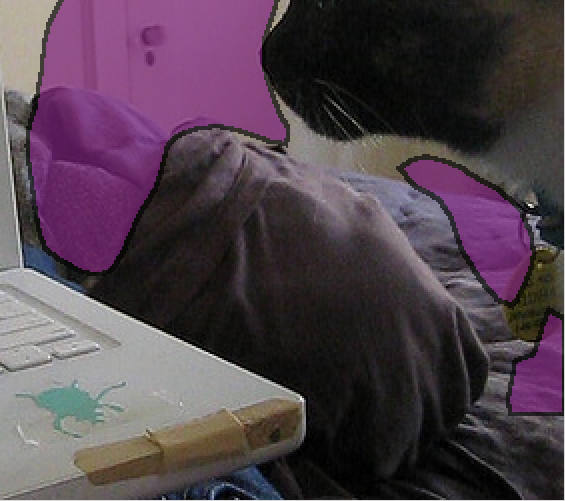}&      \includegraphics[width=0.5\columnwidth]{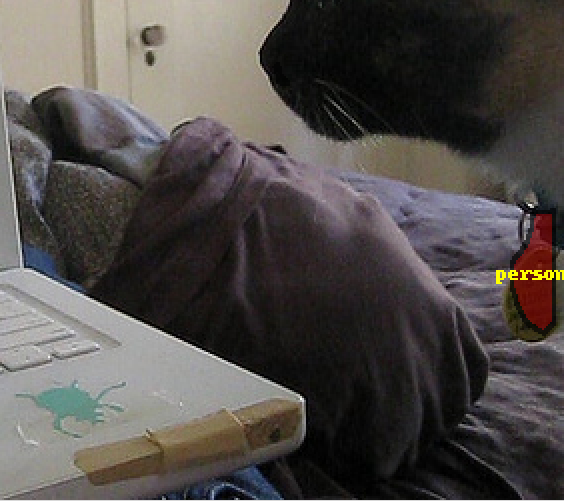}\\

    \end{tabular}
    \caption{Sequence of segments displayed to the annotator in the
      \textit{sort-by-score} setting (top row) and
      \textit{sort-by-distance} setting (bottom row). Mouse click
      position is marked with the yellow circle in the 
      image in the top-left.}
    \label{fig:ordering}
\end{figure}
\egroup




\mypar{Add segment.}
This action is initiated by left-clicking anywhere in the image.
We defined as \emph{valid} all segments in the proposal set
that contain the clicked point.
To reduce the number of redundant segments we may sort the valid segments by detection score while ignoring their class label, and then perform standard non-maximum suppression (NMS). Note that segments suppressed this way for a given add action may still be added to the annotation by clicking on another location.

We show one valid segment to the user overlaid on the current annotation. The user can scroll through the valid segments with the mouse wheel (Fig.~\ref{fig:ordering}).
We experiment with two possible orderings for scrolling: order 
by segment score or order by the Mahalanobis distance between the click location and the
segment's center of mass. Since the Mahalonobis distance is defined by
the spatial variance of the points belonging to the segment, the resulting ordering is affected by both its location and shape.
To confirm the selection of the currently visible segment, the user left-clicks a second time. This adds the selected segment to the active set and moves it in front of all other segments.

We evaluate different variants of the "Add" action in Sec.~\ref{sec:simresults}, and show that both NMS and distance-based ordering improve the efficiency of the annotation process.

\mypar{Remove segment.}
The annotator can remove any active segment by right-clicking on it.

\mypar{Change label.}
The annotator can change the label of an active segment by hovering the mouse over it and pressing any key on the keyboard. A drop-down menu appears from which the annotator can choose a label.
Instead of showing the complete label set (171 labels in our experiments), we build a shortlist of likely labels for this segment on-the-fly.
More precisely, we consider the labels of all segments that contain the current mouse position, sorted by their score.
If the correct label is not in the shortlist, the annotator can enter it manually.

\mypar{Change depth order.}
The active set of segments is ordered by depth so that each pixel is only attributed to a single segment.
The annotator can change the depth ordering of a segment by hovering the mouse over it and scrolling the mouse wheel.
Changing depth order is useful as we operate on a large set of overlapping segments, which may be in the wrong depth order. A good example is Fig.~\ref{fig:actionexample}b, where re-ordering the ``couch'' to be behind the ``person'' improves the annotation.

\mypar{Hide annotations.}
Sometimes it is hard to see the small details of the image in the interface with the segments overlaid. Pressing the space-bar temporarily hides all annotations.

\subsection{Initialization}
\label{sec:autoinit}

Fluid Annotation starts from a machine-generated initialization (Fig.~\ref{fig:actionexample}b), which we construct as follows.
We start with an active set containing only the single segment with the highest score.
Then, we add the next highest-scored segment, provided that any of its pixels are not yet covered by segments already in the active set.
We repeat this process until all segments have been considered. 
The depth ordering of the segments correspond to the order in which they have been added to the active set.

In Sec.~\ref{sec:simresults} we experimentally test whether starting Fluid Annotation from this automatic initialization is beneficial compared to starting from scratch (i.e. an empty active set).

\section{Simulator}

To efficiently explore various design options in our system we create a
simulation environment that aims to imitate the human annotation process.
For realism, the simulator operates on the very same Fluid Annotation system as the human would, including the same editing actions (Sec.~\ref{sec:actions}).

In order to mimic an annotator the simulator has access to the ground-truth full
image annotation which enables evaluating candidate actions with respect to a measure of
annotation quality.
We use the panoptic quality
metric \cite{kirillov18arxiv} as a measure for annotation quality in our
simulation. The panoptic metric is particularly well suited for our purposes
since it jointly optimizes the precision, recall and pixel-level accuracy. We
refer to \cite{kirillov18arxiv} for further details on the panoptic metric.

\mypar{Choosing between edit actions.}
Our simulator uses a greedy strategy to choose between edit actions for optimizing the
annotation quality. Before choosing an action, we generate a pool of candidate actions.
For each segment in the active set, we generate 3 candidate action:
(1) its ``remove'' action;
(2) one ``change depth order'' action by choosing the closest reordering which improves annotation quality;
(3) one ``change label'' action by setting the segment label to the label of the best matching ground-truth segment.
In addition, for each ground-truth segment which does not have a matching segment in the active set, we generate an ``add segment'' candidate action. We do this by first simulating
the mouse click and then scrolling through the set of segments available at the click location (valid segments). We stop scrolling as soon as the annotation quality improves.
Out of the pool of candidate actions, we execute the one that leads to the largest improvement in annotation quality.

\mypar{Mouse-position simulation.}  For each edit action the simulator must generate image coordinates for the mouse cursor. Note that an edit action either targets one of the segments in
the \emph{active set} (i.e. ``remove'', ``change depth order'', and ``change label''), or targets one
of the ground-truth segments (i.e. ``add segment''). To simulate positioning the mouse over a target segment we sample its position from a Gaussian distribution defined by the locations
of the pixels of the target segment. If the sampled location is not on the target
segment, we simply re-sample.

\section{Experimental results}

In this section we evaluate our annotation approach.
We first start with simulation experiments (Sec.~\ref{sec:simresults}).
Employing a simulator enables us to efficiently explore a broad range of possible settings for the interface.
In the second batch of experiments we evaluate the performance of Fluid Annotation when operated by human annotators, in the best settings found during the simulations (Sec.~\ref{sec:humanresults}).

\mypar{Evaluation metrics.}
To perform one edit action in our system the annotator has to
perform several interactions with the GUI. We denote these as
\textit{micro-actions}.
For example, ``remove'' amounts to a single micro-action corresponding to a mouse
click on a segment, whereas the ``add'' action is composed of a click on a
new location, several scrolls of the mouse wheel to sweep through the candidate
segments, and another click to confirm the selection of the current candidate.
To evaluate the effectiveness of the annotation process we measure the quality of the annotation as a function of the number of \mact spent to reach a that level of quality (averaged over all images).
We measure annotation quality by recall at IoU $>0.5$.
For thing classes, each object instance contributes separately to recall.
For each stuff class we measure IoU by treating all pixels of that class in an image as a single region (both in the ground-truth and in the output of our method). This matches the panoptic metric~\cite{kirillov18arxiv}.
For the human annotation experiments, we also include
time measurements in seconds,
and we measure agreement between multiple annotators by the average pixel accuracy (as done by~\cite{caesar18cvpr,cordts16cvpr,zhou17cvpr}).

\mypar{Dataset.}  
We evaluate our interface on the COCO 2017 validation set (5K images).
We use the ground-truth annotation provided by~\cite{lin14eccv, caesar18cvpr} that
includes 80 thing classes~\cite{lin14eccv} and 91 stuff
classes~\cite{caesar18cvpr}.
This data is highly complete: 94\% of all pixels are annotated in the ground-truth~\cite{caesar18cvpr}.

We randomly split the validation set into 
500 images that we use to explore various settings of the interface,
and a hold-out set of 4500 images on which we perform the final evaluation of the
best performing settings.
To evaluate performance with human annotators we use smaller sets of
20 and 25 randomly sampled images from our hold-out set.

We train our segmentation model (Mask-RCNN) on the COCO 2017 challenge training sets~\cite{caesar18cvpr,lin14eccv}.
We train one model on the object detection challenge training set
(120k images with 80 thing classes), and a second model on the
stuff challenge training set (40k images with 91 stuff classes). 
These training sets do not overlap with the validation set.

\subsection{Results using simulations}
\label{sec:simresults}
\begin{figure}[t]
    \centering
    \begin{tabular}{c}
      \includegraphics[width=0.95\columnwidth]{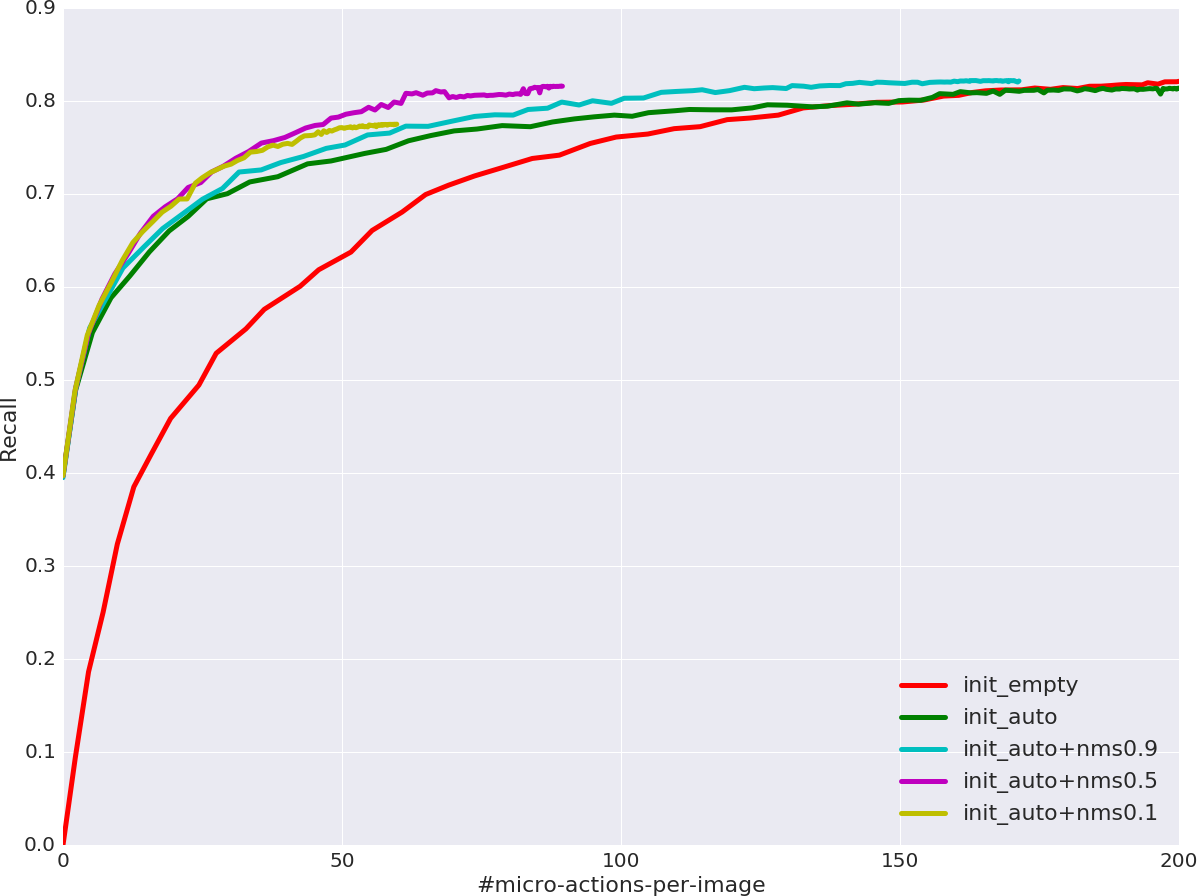}\\
    \end{tabular}
    \caption{Performance using the basic settings of our system with and without automatic initialization, and comparison for different NMS thresholds.}
    \label{fig:basicmodel}
\end{figure}

We first evaluate our Fluid Annotation interface using its basic settings: for the ``add'' action we do not apply NMS and we order the segments by their score (Sec.~\ref{sec:actions}).
We consider both a machine-generated initialization (\initauto) and starting from scratch (\initempty) (Sec.~\ref{sec:autoinit}).
Intuitively, a good initialization should save annotation time. A bad initialization would require many corrections and it may increase time instead.

Results are shown in Fig.~\ref{fig:basicmodel}.
First of all, we observe that recall rapidly increases during the first few micro-actions: using only 50 micro-actions one reaches 63\% (\initempty) or 74\% (\initauto) recall. This demonstrates that, even in its basic settings, our interface is very effective especially in the beginning of the annotation process.

Second, the machine-generated initialization kickstarts the annotation process already at 40\% recall, and this is more than doubled to 83\% after 200 micro-actions.
Example initialization mistakes are shown in Fig.~\ref{fig:actionexample} (b): The ``door'' is mistaken as part of the ``wall'', and the ``blanket'' is labeled as ``person'', while the ``keyboard'' and ``metal'' medal of the cat were missed altogether. This demonstrates that our machine segmentation model has still plenty to learn and can benefit from more training data. 

Finally, \initauto leads to a substantially better recall curve compared to \initempty: after 50 micro-actions it results in 74\% recall whereas \initempty only leads to 63\% recall.
This shows that the machine-generated initialization is clearly beneficial and we adopt this in all subsequent experiments.

\mypar{Statistics of actions and micro-actions.}
\begin{figure}[t]
    \centering
    \begin{tabular}{c}
    \includegraphics[width=0.9\columnwidth]{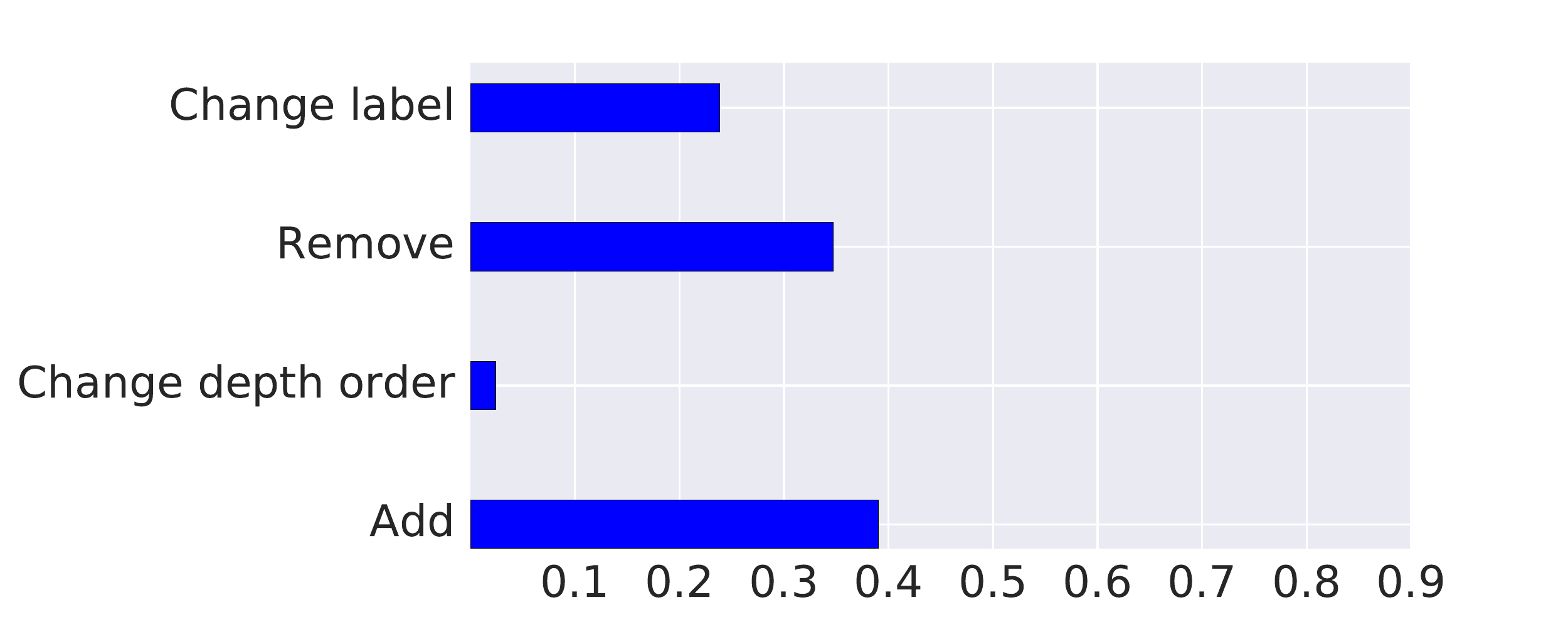}\\
      (a) Distribution of action types\vspace{0.2cm}\\
    \includegraphics[width=0.9\columnwidth]{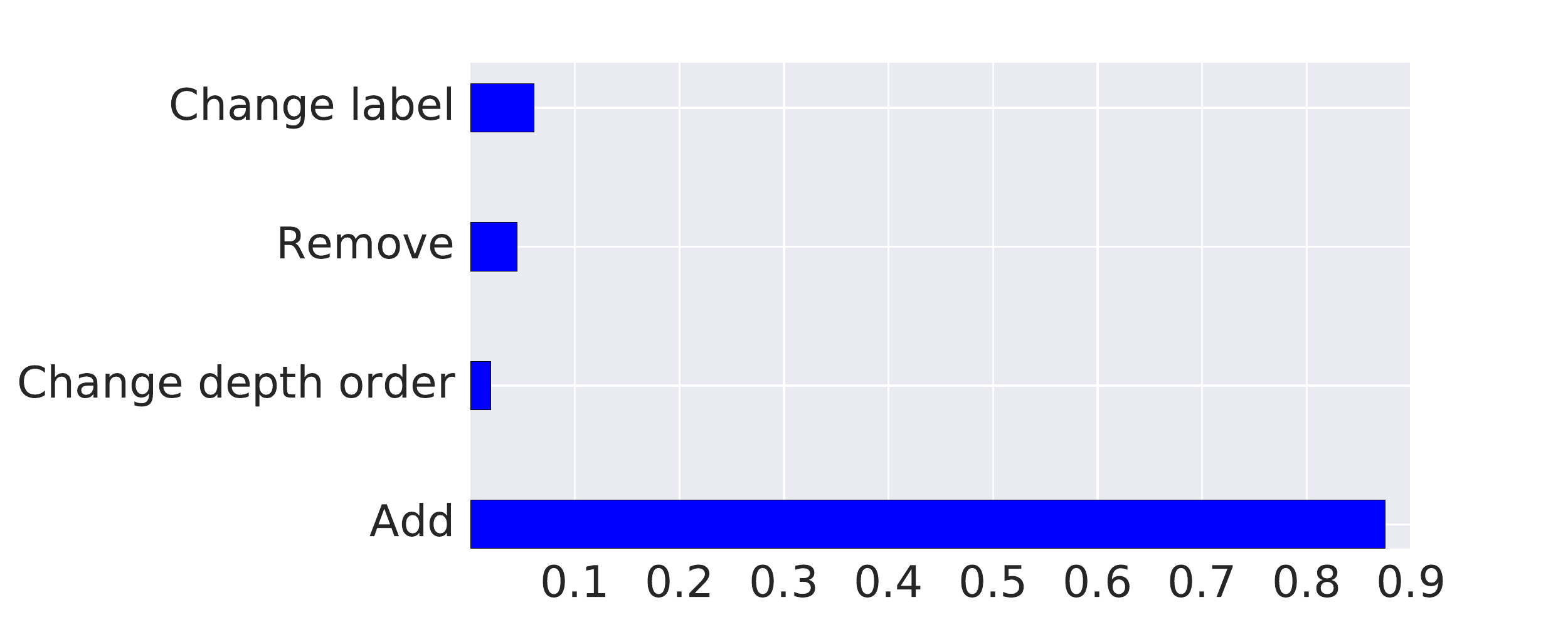}\\
      (b) Distribution of micro-action for each action type\vspace{0.2cm}\\
    \includegraphics[width=0.9\columnwidth]{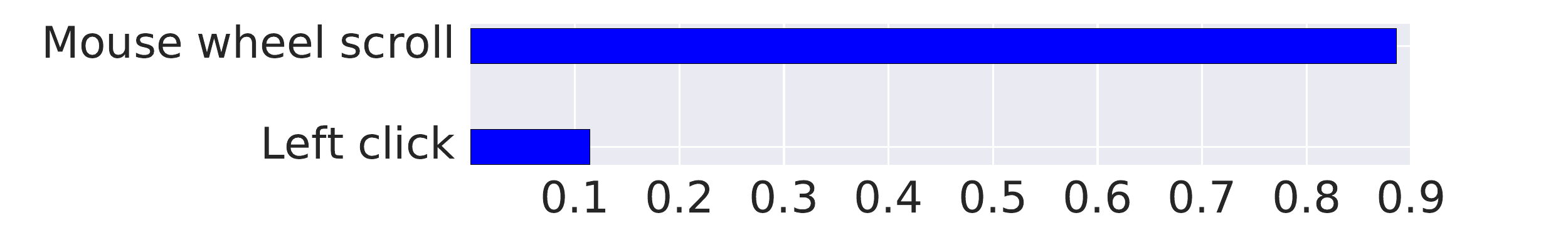}\\
      (c) Distribution of micro-actions within ``Add'' action\\
    \end{tabular}
    \caption{Distribution of actions and micro-actions.}
    \label{fig:actionstats}
\vspace{-.5cm}
\end{figure}

To identify directions for improving over the basic settings we examine
the distributions of actions and micro-actions performed during annotation (always using \initauto).
In Fig.~\ref{fig:actionstats} we show (a) the proportions of different action
types performed, 
(b) the distribution of \mact load per action type,
and (c) the distribution of micro-actions for action "Add".
Note how the lion share of \mact is consumed by the ``Add'' action, and that most of these \mact are mouse wheel ``scroll'' micro-actions used to select the best-fitting segment at the clicked location. The next experiments therefore focus on this segment selection process.

\mypar{Reducing redundancy with NMS.}
We apply NMS on the valid set of segments to reduce redundancy (Sec.~\ref{sec:actions}).
Results for various NMS thresholds are shown in Fig.~\ref{fig:basicmodel}.
We see that for a small threshold of $0.1$ too many segments are removed,
hurting the overall recall.
On the other extreme, a high threshold of $0.9$ does preserve recall, but removes only a few segments.
A moderate NMS threshold of 0.5 instead allows to preserve recall
while substantially reducing the number of segments that need to
be considered during the "Add" action. At this threshold, the
top recall
is reached after just 61 micro-actions, compared to more than 150 \mact without NMS.
We therefore adopt \initautonms in all subsequent experiments.

\begin{figure}[t]
    \centering
    \begin{tabular}{cc}
      \includegraphics[width=0.45\columnwidth]{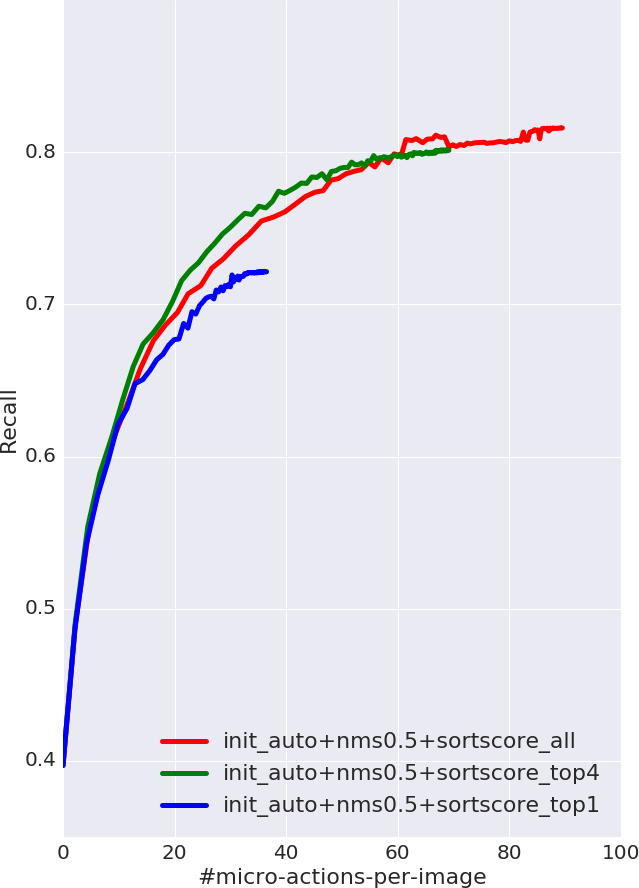} &
      \includegraphics[width=0.45\columnwidth]{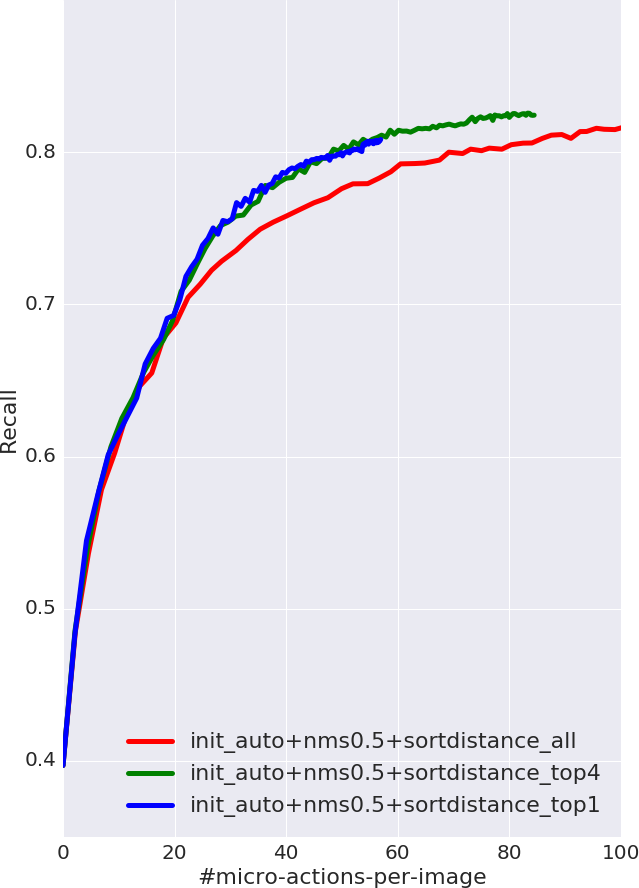}\\
      (a) & (b)\\
    \end{tabular}
    \caption{Ordering of segments according to Mask-RCNN score (a) vs. Mahalanobis distance between the click location and the segment center (b).}
    \label{fig:segselect}
\vspace{-0.3cm}
\end{figure}

\mypar{Sorting by score vs by distance.}  
In Fig.~\ref{fig:segselect} we compare the two segment orderings introduced in Sec.~\ref{sec:actions} (i.e. by segment score, or by the Mahalanobis distance between the click and the segment's center).
To highlight the ability of each ordering to place the correct segment early we limit the maximum number of segments available to annotator during the "Add" action.
We denote our system settings with limit to $N$ top segments and ordering by
score as \initautonmsSSN, and settings with ordering by distance
as \initautonmsSDN.

Without top $N$ limiting the curves for both orderings are about the same and achieve 82\% recall in 90 micro-actions (compare the red curves in Fig.~\ref{fig:segselect} (a) and (b)). When setting the limit to top $4$, distance ordering achieves a higher final recall than score ordering, 83\% vs 80\% (compare green curves). It is also more efficient: at 40 \mact, it yields 79\% recall instead of 77\%. We conclude that distance ordering and using top 4 is the most efficient settings, and it does not reduce final recall. We use this for subsequent experiments.

To better understand this result, consider the extreme case of limiting to the top $1$ segment only. When the annotator clicks near the center of a missing object, the top 1 segment ordered by distance will be the right one (assuming it is in the proposal set).
In contrast, when ordering by score, the top 1 segment is potentially unrelated to the missing object. In the worst case, the overall highest scored segment might occupy the whole image. In that case, it will be the only available segment, regardless of where the annotator clicks.
\begin{figure}[h!]
    \centering
    \begin{tabular}{c}
      \includegraphics[width=0.95\columnwidth]{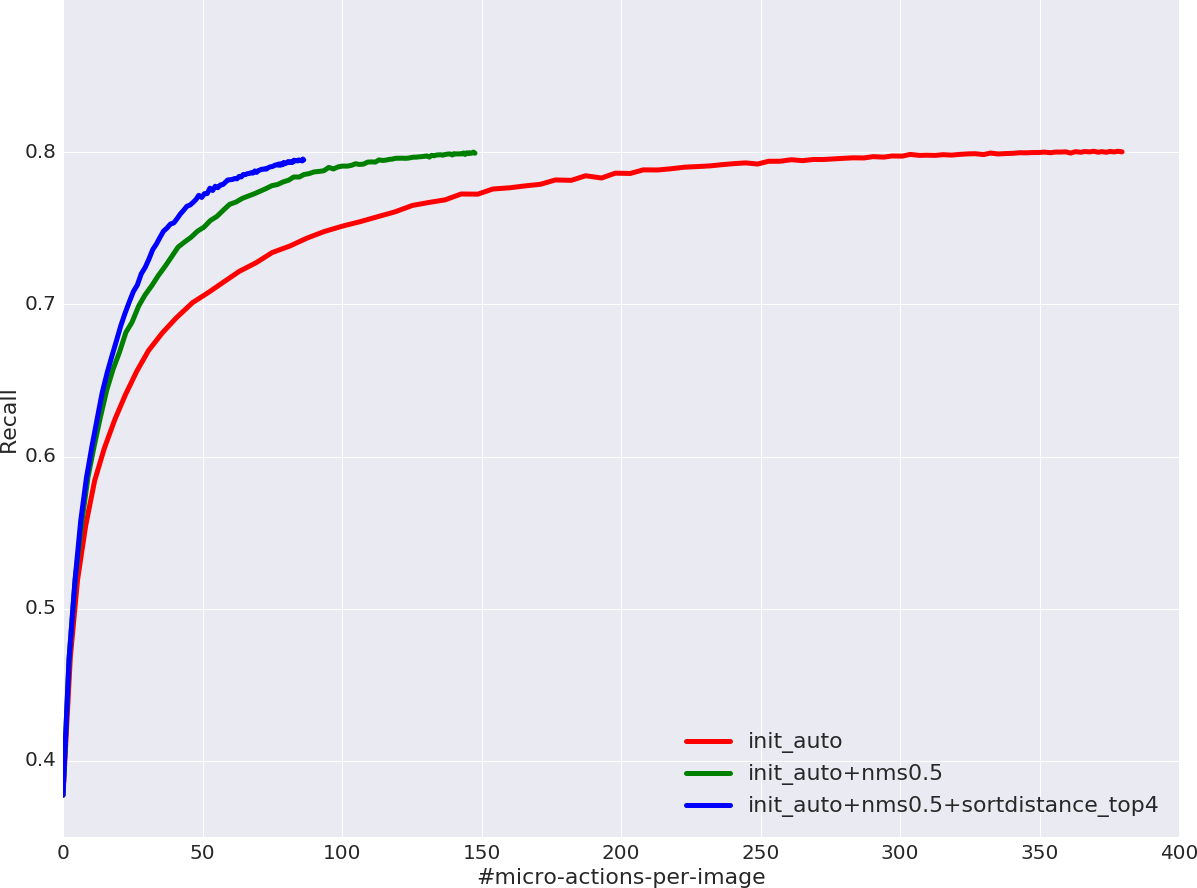}
    \end{tabular}
    \caption{Comparison of our basic settings (red), using NMS within ``Add'' (green), and using the best settings (blue) on the hold-out set of 4500 images.}
    \label{fig:best}
\end{figure}


\mypar{Validation of results on hold-out set.}
We verify the effectiveness of the chosen settings on the hold-out set of 4500 images.
In particular, Fig.~\ref{fig:best} compares \initauto, \initautonms, and \initautonmsSDT.
We observe that the improvements hold: while in all experiments recall reaches 80\%, we improve the number of micro-actions necessary to get there improve from 350 for the basic settings (\initauto), to 147 when using NMS during "Add", and to just 86 when also sorting by distance and limiting selection to the top $4$ segments.

\subsection{Results with human annotators}
\label{sec:humanresults}

\begin{table}[t!]
\small
    \centering
    \begin{tabular}{|c|c|c|}
    \hline
    \multicolumn{3}{|c|}{\textbf{Human pixel-wise label agreement}} \\
    \hline
    Fluid annotation vs. & Fluid Annotation vs. & Polygon annotation vs. \\
    COCO+stuff original & Polygon annotation & COCO+stuff original \\
    \hline
    69\% & 66\% & 65\% \\ 
    \hline
    \end{tabular}
    \caption{Pixel-wise label agreement across three different annotation methods.}
    \label{tab:human_pixel_agreement}
    \begin{tabular}{|c|c|c|}
    \hline
    \multicolumn{3}{|c|}{\textbf{Average annotation time per image}} \\
    \hline
     Fluid annotation & Polygons & COCO+stuff original~\cite{lin14eccv,caesar18cvpr}\\
    \hline
     175 & 507 & $\approx$ 1140\\
    \hline
    \end{tabular}
    \caption{Average annotation time for human experts using the Fluid Annotation interface and the polygon-based interface of LabelMe, and for crowdsourced annotators on COCO+Stuff (16 minutes in~\cite{lin14eccv} plus 3 minutes in~\cite{caesar18cvpr}.)}
    \label{tab:annotation_time}
\end{table}

We now perform several experiments with expert human annotators using the best settings determined in Sec.~\ref{sec:simresults}: use the machine-generated initialization, and, for the ``Add'' action, use NMS with IoU $>0.5$, sort by distance to segment center, and limit selection to the top $4$ segments.

\mypar{Reproducing a reference annotation.} 
We first verify how a human annotator can reproduce a reference annotation, which tests both the flexibility and efficiency of the annotation interface.
To do this, we ask two human experts to annotate 25 images while looking at their reference annotation from~\cite{caesar18cvpr,lin14eccv} using two interfaces:
(I) Fluid Annotation,
and
(II) A polygon-based interface representative for what was used to annotate many datasets~\cite{cordts16cvpr,mottaghi14cvpr,xiao2014sun,zhou17cvpr}. More precisely, we use the popular LabelMe~\cite{russell08ijcv} as implemented in~\cite{labelmepython}.

We measure quality in terms of recall and efficiency in terms of \mact, which allows comparing to our simulation results.
For the LabelMe interface~\cite{russel08ijcv}, each polygon the annotator draws is consider to cost $p+2$ \mact, where $p$ is the number of polygon vertices, plus 2 to input its label (type name and confirm). 

\begin{figure}[t]
    \centering
    \begin{tabular}{c}
      \includegraphics[width=0.95\columnwidth]{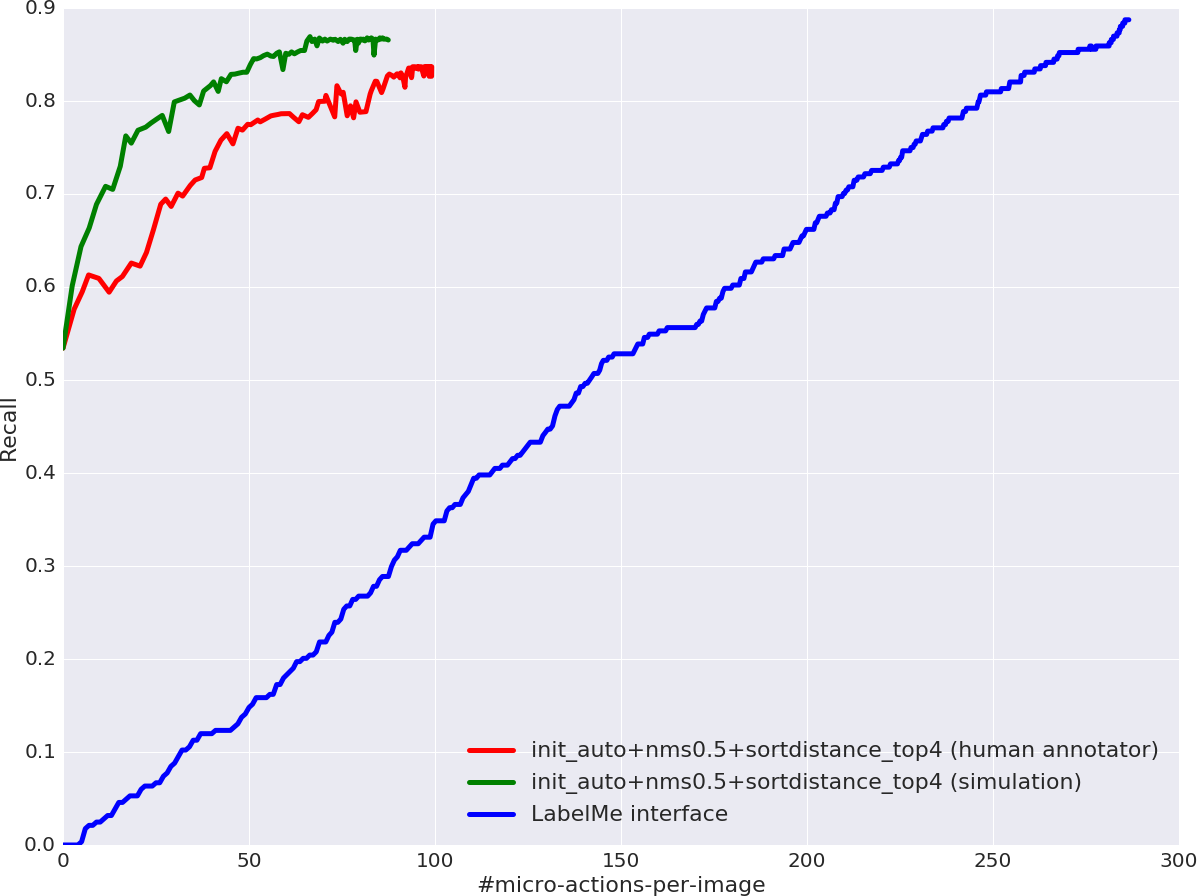}\\
    \end{tabular}
    \caption{Comparison of annotation results obtained with our Fluid Annotation system (green, red) and with the LabelMe interface (blue) \cite{russel08ijcv,labelmepython}.}
    \label{fig:human_results}
\vspace{-0.4cm}
\end{figure}

Results are shown in Fig.~\ref{fig:human_results}.
The performance curve for the human annotator behave in a very similar manner to the simulator one.  As noted before for the simulator, the recall produced by the human grows rapidly, especially during the first few micro-actions. Moreover, the simulator needs 68 \mact per image to produce a recall of 87\%, while the human needs 96 \mact to get to a slightly lower recall (83\%).
This similarity in behaviour shows that the simulator is a good proxy for human performance, and that Fluid Annotation is an interface a human can truly efficiently operate.

Importantly, Fluid Annotation is substantially more efficient than the LabelMe interface: using 100 \mact, a human annotator produces 35\% recall with LabelMe while 83\% with Fluid Annotation. From another view, to reach 83\% recall, LabelMe requires 2.5$\times$ more \mact. This demonstrates that Fluid Annotation is a highly effective interface.

\mypar{Human agreement for different interfaces.}
While so far we have assumed that the reference annotations~\cite{caesar18cvpr,lin14eccv} are perfect ground-truth, in practice different humans annotating the same image typically produce somewhat different annotations. Therefore, often researchers measure the agreement across multiple annotators~\cite{cordts16cvpr,caesar18cvpr,zhou17cvpr}.
To do this, we annotate 20 images with two expert annotators, this time by showing just the image, without any reference annotation.
The first 10 images are annotated by annotator A using Fluid Annotation, and by annotator B using the LabelMe interface.
For the second 10 images the annotators switch interfaces: A uses LabelMe and B uses Fluid Annotation. This protocol removes any possible annotator bias from the comparison between interfaces. We measure annotation time and pixel-wise label agreement~\cite{cordts16cvpr,caesar18cvpr,zhou17cvpr} of all three forms of annotation:
Fluid annotation,
LabelMe (polygons using our expert annotators),
and the process of COCO+stuff~\cite{caesar18cvpr,lin14eccv} (polygons for thing classes and superpixel annotation for stuff classes, all using crowdsourced annotators).


\begin{figure*}[t]
\vspace{-.0cm}
    \begin{tabular}{cccc}
      original image & COCO+Stuff & LabelMe & Fluid Annotation \\
      \includegraphics[width=0.24\textwidth]{annotation_examples/000608_162732.png}&
      \includegraphics[width=0.24\textwidth]{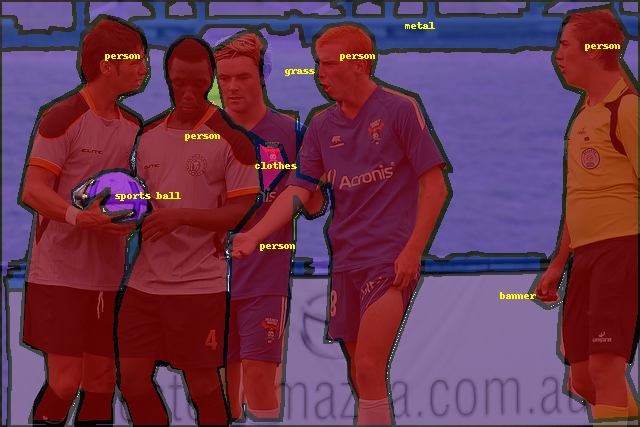}&
      \includegraphics[width=0.24\textwidth]{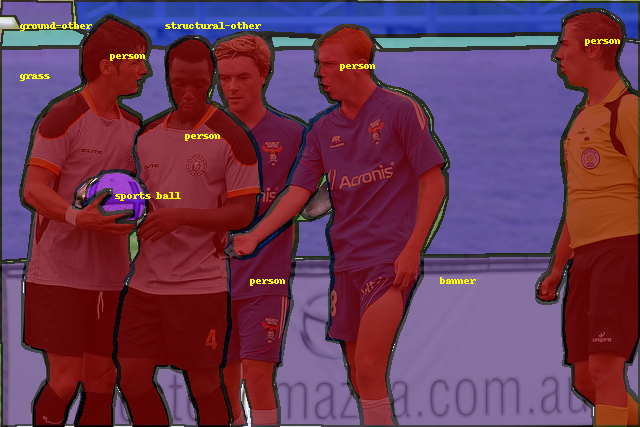}&
      \includegraphics[width=0.24\textwidth]{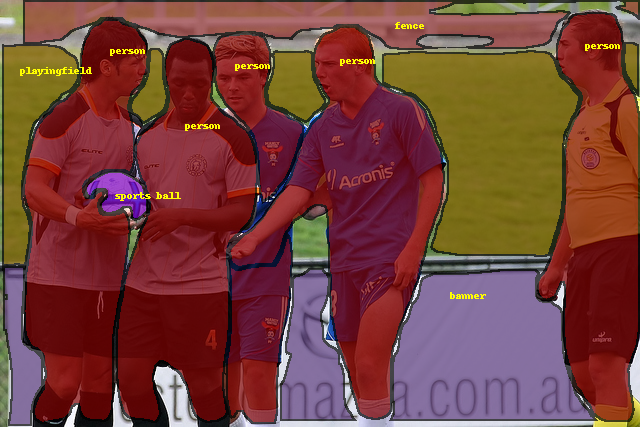}\\
    \end{tabular}
    \begin{tabular}{cccc}
      \includegraphics[width=0.24\textwidth]{annotation_examples/000657_166747.png}&
      \includegraphics[width=0.24\textwidth]{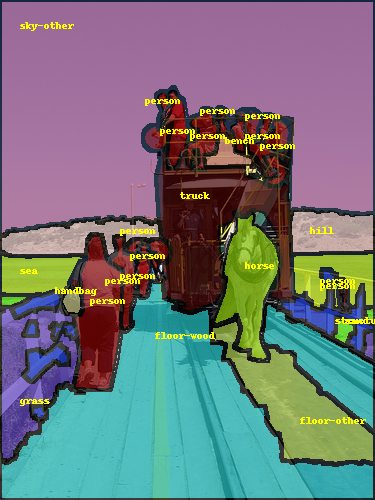}&
      \includegraphics[width=0.24\textwidth]{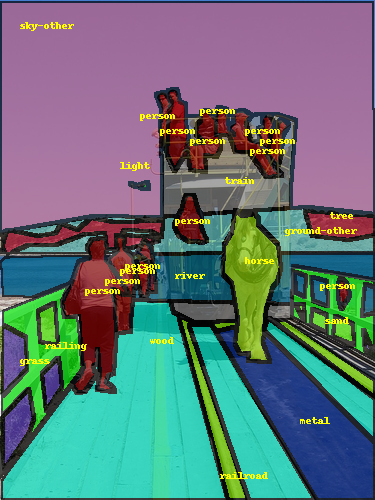}&
      \includegraphics[width=0.24\textwidth]{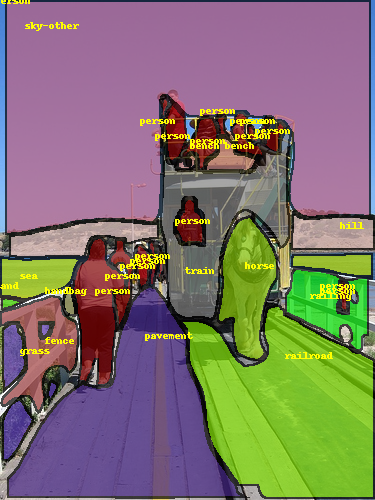}\\
    \end{tabular}
    \begin{tabular}{cccc}
      \includegraphics[width=0.24\textwidth]{annotation_examples/000653_166521.png}&
      \includegraphics[width=0.24\textwidth]{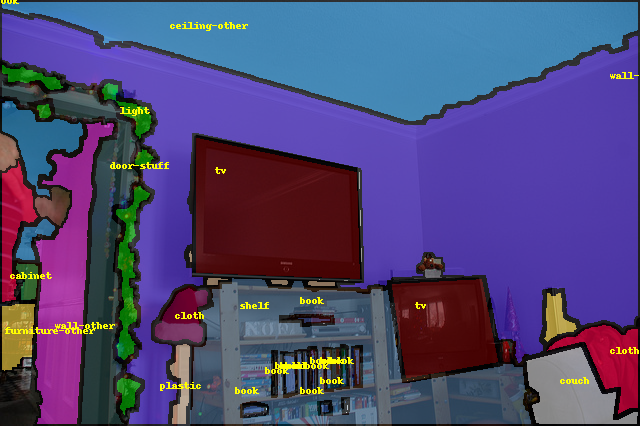}&
      \includegraphics[width=0.24\textwidth]{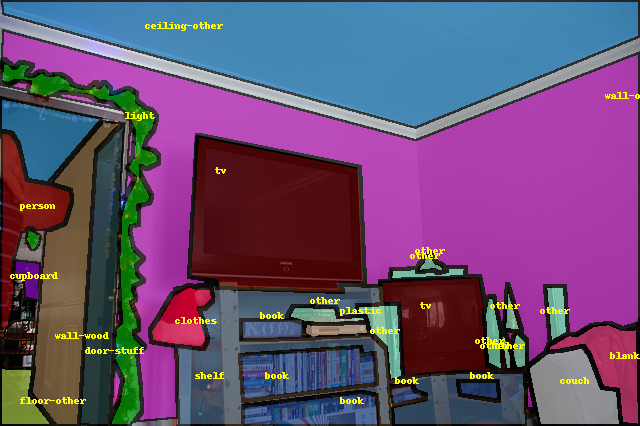}&
      \includegraphics[width=0.24\textwidth]{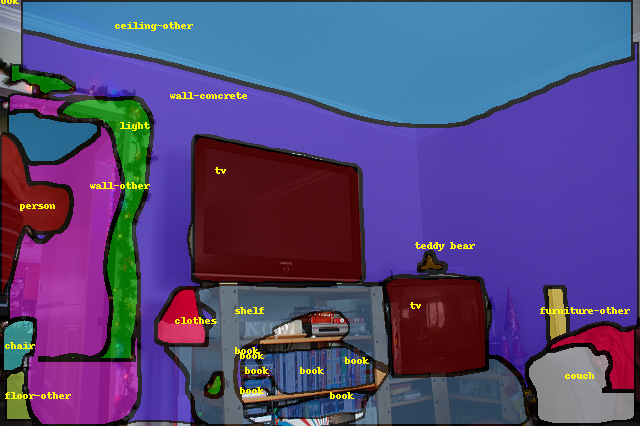}\\
    \end{tabular}
    \caption{Comparison of annotations created using different interfaces: The original COCO+stuff annotations~\cite{caesar18cvpr,lin14eccv}, LabelMe~\cite{russel08ijcv} polygon annotations, and our Fluid Annotations.}
    \label{fig:annotation_examples}
\end{figure*}

The results are presented in Tab.~\ref{tab:human_pixel_agreement} and~\ref{tab:annotation_time}.
All label agreements are relatively close, ranging from 65\% to 69\%. 
This level of agreement is reasonable.
COCO-stuff~\cite{caesar18cvpr} reports 74\% label agreement on stuff only, where fewer classes can be confused.
The authors of the ADE20 dataset~\cite{zhou17cvpr}, which was annotated using LabelMe, report 82\% agreement using the same expert annotator six months apart, while 33\% agreement between different expert annotators.
Therefore, we conclude that the quality of annotations for fluid annotation is on par with the compared methods.

\mypar{Annotation time.}
Table~\ref{tab:annotation_time} compares annotation time across annotation interfaces. We see that Fluid Annotation is $3\times$ faster than the LabelMe interface (Table~\ref{tab:annotation_time}).
In turn, our annotators with the LabelMe interface are twice as fast as what originally reported for COCO+stuff~\cite{caesar18cvpr,lin14eccv}, despite that the bulk of their annotation time was also consumed by drawing polygons. However, this can be attributed to using crowdsourcing versus human experts.

Looking within Fluid Annotation, the action frequencies of an annotator can be broken down as follows: 33\% ``add'', 31\% ``remove'', 26\% ``change label'', and 10\% ``change depth order''.
The ``add'' action is the most expensive action and takes 18\% of the total annotation time. Interestingly, 66\% of the total time is spent \emph{between} actions, when the annotator observes the image and decides what to do next.

\mypar{Qualitative examples.}
Fig.~\ref{fig:annotation_examples} shows qualitative examples of the various annotation strategies.
Generally, Fluid Annotation yields good outlines for most objects.
When comparing annotations made by different interfaces, we observe that most of the disagreements are caused by similar segments having a slightly different label.
For example, the wall of the third image is sometimes annotated as
\emph{wall-concrete} and sometimes as \emph{wall-other}. As another example,
there is legitimate disagreement about how exactly the wooden walk-boards in the
second image should be labeled. Finally, some cases are very difficult for Fluid
Annotation: in the third image, everything in the doorway on the left is blurry,
while the Christmas garland around the door frame is very irregular. As such,
the produced segments are only roughly following the actual object
boundaries. In fact in these image conditions typically even polygon or
superpixel interfaces would produce only approximate outlines.

\section{Conclusion}
We presented Fluid Annotation, an intuitive human-machine collaboration interface for annotating the class label and outline of every object and background region in an image.
Fluid annotation substantially reduce human annotation effort,
supports full images annotation in a single pass, and it empowers the annotator to choose what to annotate and in which order.
We have experimentally demonstrated that Fluid Annotation takes $3\times$ less annotation time than the popular LabelMe interface.

\bibliographystyle{ACM-Reference-Format}
\bibliography{shortstrings,loco}

\end{document}